# Non-linear Visual Knowledge Discovery with Elliptic Paired Coordinates

Rose McDonald, Boris Kovalerchuk
*Dept. of Computer Science*
*Central Washington University, USA*
Rose.McDonald@cwu.edu, Borisk@cwu.edu

**Abstract**. It is challenging for humans to enable visual knowledge discovery in data with more than 2-3 dimensions with a naked eye. This chapter explores the efficiency of discovering predictive machine learning models interactively using new Elliptic Paired coordinates (EPC) visualizations. It is shown that EPC are capable to visualize multidimensional data and support visual machine learning with preservation of multidimensional information in 2-D. Relative to parallel and radial coordinates, EPC visualization requires only a half of the visual elements for each n-D point. An interactive software system EllipseVis, which is developed in this work, processes high-dimensional datasets, creates EPC visualizations, and produces predictive classification models by discovering dominance rules in EPC. By using interactive and automatic processes it discovers zones in EPC with a high dominance of a single class. The EPC methodology has been successful in discovering non-linear predictive models with high coverage and precision in the computational experiments. This can benefit multiple domains by producing visually appealing dominance rules. This chapter presents results of successful testing the EPC non-linear methodology in experiments using real and simulated data, EPC generalized to the Dynamic Elliptic Paired Coordinates (DEPC), incorporation of the weights of coordinates to optimize the visual discovery, introduction of an alternative EPC design and introduction of the concept of incompact machine learning methodology based on EPC/DEPC.

## I. Introduction

The efficient use of visualization in Machine Learning (ML) requires preservation of multidimensional information and **Elliptic Paired Coordinates (EPCs)** is one of the visualization methods that preserves this n-D information [1,10].

Commonly *n-D points* are mapped *to 2-D points* for visualization, which *can approximate n-D information partially preserving it* in the form of *similarities* between n-D points by using MDS, SOM, t-SNE, PCA and other *unsupervised* methods [6, 11,12, 14]. Moreover, both n-D similarities injected by these methods and their 2-D projections for visualization can be irrelevant to a given learning task [1]. In [12] t-SNE is used to visualize not only the input data but also the activation (output) of neurons of a selected hidden layer of the MLP and CNN trained models. This combines unsupervised t-SNE method with the information learned at this layer. However, projection of the multidimensional activation information of the layer to 2-D for visualization is lossy. It only partially preserves the information of that layer. Moreover, the layer itself only partially preserves n-D input information. Thus, some information that is important for classification can be missed.

An alternative methodology is *mapping n-D points to 2-D graphs* that preserves all n-D information [1,2,5,13]. Elliptic Paired Coordinates belong to the later. In both methodologies the respective visual representations of n-D data are used to solve predictive classification tasks [7].

The advantages of EPC include: (1) *preserving* all n-D information, (2) capturing *non-linear dependencies* in the data, and (3) requiring fewer visual elements than methods such as parallel and radial coordinates.

This chapter expands our prior work [10] on EPC in: (1) successful *testing* the EPC approach and methodology in additional experiments with both real and simulated data, (2) generalizing EPC to the *Dynamic* Elliptic Paired Coordinates (DEPC), (3) generalizing EPC and DEPC by incorporating the *weights* of coordinates to optimize the visual discovery in EPC/DEPC, alternating side ellipses and introducing the concept of *incompact* machine learning methodology based on EPC/DEPC.

The chapter is organized as follows. Section 2 describes the concept of elliptic paired coordinates. Section 3 presents a visual knowledge discovery system based on the elliptic paired coordinates. Section 4 describes the results of experiments with multiple real data sets. Section 5 presents experiments with synthetic data. Section 6 provides generalization options for elliptic paired coordinates and Section 7 concludes the paper with its summary and future work.

2. ELLIPTIC PAIRED COORDINATES

*2.1 Concept*

In EPC coordinate axes are located on ellipses (see Fig. 1). In [1] the EPC shows in Fig. 1 is called EPC-H, in this chapter we omit H. In Fig. 1, a short green arrow losslessly represents a 4-D point P = (0.3,0.5,0.5,0.2) in EPC, i.e., this 4-D point can be restored from it. The number of nodes in EPC is two times less than in parallel and radial coordinates with less occlusion of lines. For comparison see point P in Fig. 2 in radial and parallel coordinates with 4 nodes instead of 2 nodes in EPC. The dark blue ellipse $C_E$ in Fig. 1 contain four coordinate curves $X_1$-$X_4$. It is called the **central ellipse.**

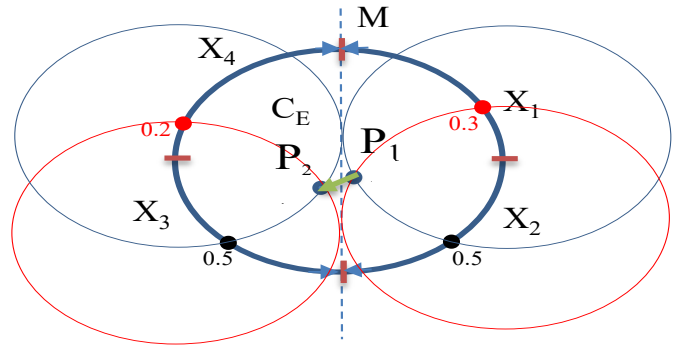

Fig. 1. 4-D point P = (0.3,0.5,0.5,0.2) in 4-D EPC as green arrow $P_1{\rightarrow}P_2$. Red marks separate coordinates in the blue coordinate ellipse.

Fig. 1 shows only one of the options how we can split an ellipse or a circle to the segments. Here each coordinate starts at the horizontal red marks on the right or on the left edges of the central ellipse, where $X_1$, $X_4$ go up and $X_2$, $X_3$ go down from respective points. An alternative *sequential way* of directing coordinates is that $X_1$ starts from the top, ends in the right middle point, where $X_2$ starts and go down to the bottom point, where $X_3$ starts as so on. This is a simpler way when we have more than 4 attributes to be mapped to the ellipse. In general, the direction of each coordinate can be completely independent of the directions of the other coordinates. While all these possibilities exist for EPC, we use either one which is shown in Fig.1 or a sequential order from the top of the central ellipse. Another option called Dynamic EPC is present later in section 6.1.

Four **side ellipses** of the size of the central blue ellipse are used to build the green arrow from point $P_1$ to point $P_2$, $P_1 \rightarrow P_2$. The middle vertical line M is guiding line for side ellipses. They touch line M. Moving these side ellipses along line M produces different 4-D points. The thin red side ellipse on the right goes through $x_1 = 0.3$, touch line M and has the size of the central ellipse $C_E$.

The thin blue side ellipse on the right is built in the say way for $x_2 = 0.5$. The point $P_1$ is the **crossing point** where these red and blue side ellipses cross each other in $C_E$. The point $P_1$ represents pair $(x_1,x_2) = (0.3,0.5)$. The point $P_2$ that represents pair $(x_3,x_4) = (0.5,0.2)$ is constructed in the same way for $x_3 = 0.5$ and $x_4 = 0.2$ by generating respective red and blue side ellipses on the left. Next, the arrow from $P_1$ to $P_2$ is made (see a short green arrow in Fig. 1), which losslessly visualize 4-D point $P = (0.3,0.5,0.5,0.2)$.

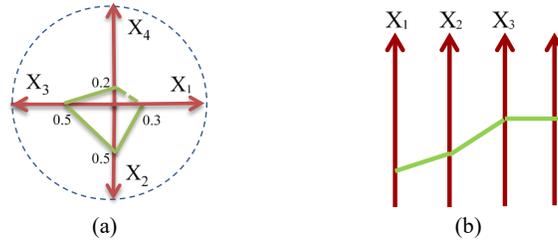

(a)             (b)

Fig. 2. 4-D point $P = (0.3,0.5,0.5,0.2)$ in radial (a) and parallel coordinates (b).

It means that we can reverse these steps to restore the 4-D point P from $(P_1 \rightarrow P_2)$ arrow. Visually we slide the red right-side ellipse to cross point $P_1$. Its crossing point with the central blue ellipse is $x_1 = 0.3$. Similarly, we get $x_2 = 0.5$ with the right-side blue ellipse. The same process allows restoring $P_2$, with red and blue left-side ellipses.

### 2.2 Elliptic Paired Coordinates Algorithm

Below we present steps of EPC algorithm for n-D data referencing Fig. 1 and notation introduced in it:
1. Create the *central ellipse*, $C_E$, with a vertical line M bisecting it.
2. Divide the circumference of $C_E$ into *n equal sectors* by the number of dimensions $X_1, X_2, \ldots, X_n$ to be represented.
3. *Normalize* all data on the scale [0,1].
4. Graph the *data points* $x_i$ along the circumference of the $C_E$ in the appropriate sector for each n-D point $\mathbf{x}=(x_1,x_2,\ldots,x_n)$.

5. Pair data points by dimension, starting at $(x_1, x_2)$. If there is an odd number of pairs of coordinates, use the horizontal bisector N for the middle pair instead of the vertical bisector M.

6. Create the red and blue side ellipses, such that they touch the line M. Have the top of red ellipses to intersect the data point $x_i$ on the central ellipse and the bottom of blue ellipses to intersect the next data point on the central ellipse and then alternate between red and blue for each coordinate in sequence.

7. Find the intersects between each pair, connect them.

*2.3. EPC mathematical formulation for n-D data*

Steps 6 of EPC algorithm requires constructing thin ellipses. Their width W and heigh H are equal to them for the central ellipse $C_E$, but it requires computing their centers. Let point (A,B) be the center of such ellipse. It can be found by soling for (A,B) the general equation for an ellipse,

$$\frac{(x-A)^2}{W^2} - \frac{(y-B)^2}{H^2} = 1$$

with given the W, H, x- and y-coordinates of the point on the ellipse. Then we can calculate the intercepts of the ellipses to form the points $P_1$ and $P_2$. Let the center of the central ellipse be the point (cx, cy) then formulas for A and B for side ellipses are as follows.

**Right and Left Ellipses**. The +/- assignment is dependent on the ellipse being to the right or left of the line M, respectively,

$$A = cx \pm W/2$$

The +/- assignment (red/blue respectively) depends on the ellipse intersecting at the top or bottom of the thin ellipse.

$$B = y \pm \sqrt{H^2\left(1 - \frac{(x-A)^2}{W^2}\right)}$$

**Top and Bottom Ellipses**. If the number of coordinates $n$ is even, but the number of pairs of coordinates $n/2$ is odd (e.g. 5 pairs from 10 coordinates), a horizontal line N across the center ellipse is used similarly to M for placing two ellipses for the selected pair $(x_i, x_j)$. These ellipses touch line N (being above or below line N) and the intersect point of these ellipses is computed. If the selected pair is $(x_{n/2}, x_{n/2+1})$, e.g. $(x_5, x_6)$ for $n=10$ then the bottom ellipse is constructed with

$$B = cy - H/2$$

If the selected pair is $(x_1, x_n)$, e.g. $(x_1, x_{10})$ for $n=10$ then then the top ellipse is constructed with

$$B = cy + H/2.$$

For the value $A$ the +/- assignment depends on the ellipse being to the right or left of the line M, respectively.

$$A = x \pm \sqrt{W^2\left(1 - \frac{(y-B)^2}{H}\right)}$$

*2.4. Elliptic Paired Coordinates pseudo-code*

The pseudo-code is as follows.

**CreateEPC**(W, H, cx, cy):
   DrawEllipse(W,H, cx,cy,)
   // Draw ellipse $C_E$ using $\frac{(x-A)^2}{W^2} - \frac{(y-B)^2}{H^2} = 1$ with (A, B) = ($c_x$, $c_y$).
   DrawDottedVerticalLineM(cx,cy) //vertical line M at point (cx,cy)
   DrawDottedHorizontallLineN(cx.cy) //horizontal line N at point (cx,cy)
   ComputeEllipseSectorLength(D) = ($C_E$ circumference) / (number of coordinates, *n)*
   If *n* is odd then *n*++ for all x(*n*), x(*n*) = x(*n*-1)
     // Duplicate the last coordinate to make *n* even.
   Normalize(x(*i*)): x(*i*) = x(*i*) – min(*i*) / (max(*i*) – min(*i*)) // normalize each x(*i*) of
     // Coordinate $X_i$, where min(*i*) and max(*i*) are min and max for coordinate $X_i$ in data.

For *i*= 1:*n*
   **DrawDataPointOnEllipse**(x(*i*))
     // Draw data point value x(*i*) along $C_E$'s axis on sector(*i*).
       If *n* is even then
         If (*i* < *n*/2) DrawRightEllipse(i) Else DrawLeftEllipse(i)
           Else if (*i* = 0 to *n*/2 – 1) DrawRightEllipse(i)
            Else if (*i* = *n*/2 – 1 to *n*/2 + 1) DrawBottomEllipse(i)
             Else DrawLeftEllipse(*i*)
   **DrawEPC_Point**(*i*,*i*+1)) // For each pair ($x_1,x_2$), ($x_3,x_4$),…,($x_i,x_{i+1}$), ($x_{n-1},x_n$)
   // Compute a crossing point $P_{i,i+1}$ of two ellipses.
     If *n* is odd then
       If (*i* < n/2) DrawTopEllipse(i) Else DrawBottomEllipse(i)
         // For each data point dimension pair $P_i$, draw $P_{i.1}$ to $P_{i.2}$.

**DrawGraph**() // Draw arrows to connect points $P_{1,2}$, $P_{3,4}$,…, $P_{i,i+1}$,…, $P_{n-1,n}$.

3 EPC VISUAL KNOWLEDGE DISCOVERY SYSTEM

*3.1 Dominance Rectangular Rules (DR2) algorithm in EPC*

Section 2 described how n-D data points are visualized in EPC losslessly. This section presents the algorithm that discovers Dominance Rectangular Rules (DR2) based on interactive or automatic finding the rectangular areas where a single class dominates other classes. In EPC an n-D point **x**=($x_1,x_2,…,x_n$) is represented as a graph x* , where each node encodes a pair of values ($x_i,x_{i+1}$).

Consider a set of n-D point visualized in EPC, a class Q and a rectangle R in the EPC, then the forms of the rule **r** that we explore in this chapter are:
**Point Rule r**: If a node of graph **x*** of n-D point **x** is *in* R then **x** is in class Q.
**Intersect Rule r**: If graph **x*** of n-D point **x** *intersects* R then **x** is in class Q.

A point rule captures a non-linear relation of *2 attributes* that are encoded in the graph node in R. An intersect rule captures a non-linear relation of *4 attributes* that form a line that crosses R. This line is the edge that connects two nodes of the graph **x***.

We discover both types of rules using the DDR algorithm. The steps of the **DRR algorithm** are:
1. Visualize data in EPC;
2. Set up parameters of the dominant rectangles: coverage, dominance precision thresholds and type (intersect or point);
3. Search for dominant rectangles:
3.1. Automatic search: setting up the size of the rectangle, shift rectangle with a given step over the EPC display, compute coverage and precision for each rectangle, record rectangles that satisfy thresholds;
3.2. Interactive search: draw a rectangle of any size and at any location in the EPC display, collect coverage and precision for this rectangle;
4. Remove/hide cases that are in the accepted rules and continue search for new rules if desired.

The automatic search involves the *dominance search parameters*: coverage (recall in the class) and precision thresholds. A user gets only rules that satisfy these thresholds. We typically used at least 10% coverage and 90% precision.

*3.2 EllipseVis: Interactive Software System*

In **EllipseVis** users can interactively conduct visual analysis of datasets in EPC and discover classification rules. EllipseVis is available at GitHub [3]. A user can more easily detect patterns to classify data by moving the camera, and hiding various elements of the visualization.

In the automatic mode EllipseVis can do calculations to find classification patterns. EllipseVis checks if most data points within the rectangle belongs to a single class when dividing the EPC space into smaller rectangular sections. We call such rectangles *class dominant rectangles* or shorter as dominant rectangles. Respectively rules that correspond to dominance rectangles discovered by EllipseVis are called dominance rules. The quality of rules is defined by the total number of points in the rectangle (rule coverage) and the percentage of dominant class compared to the others (precision of classification). In the interactive mode a user creates the dominance rules in EllipseVis. After a rule is created, the data points that satisfy the rule, i.e., inside the rectangle, are removed from the rest of the data and further rules are discovered without them.

Several interactive capabilities are implemented in EllipseVis such as:
- Camera move (pan and zoom)
- Setting of a dominance rectangle: minimum coverage and precision; rectangle dimensions;
- Toggle whether the rectangles are calculating data on their points or intersect lines; Automatically find dominance rectangle rules, Clear rectangles.
- Hide or show elements: intersect lines, dominance rectangles, cycle between showing all lines, lines not within rules, and line within rules, Side, thin red and thin blue ellipses.

## 4 EXPERIMENTS WITH REAL DATA

The goal of experimentation is exploring efficiency of EPC on several benchmark datasets from UCI ML repository [4].

### 4.1 Experiment with Iris data

All 150 cases of Iris 4-D data of 3 classes are shown in Fig. 3 losslessly in EPC with a small insert showing them inside of the 4-D elliptic coordinates. Discovered classification rules **r₁-r₃** are represented by rectangles $R_1$-$R_3$ in a zoomed part in Fig. 3. Two misclassified red cases are presented zoomed in Fig. 4. The comparison with Iris data in parallel coordinates shown in Fig. 5 demonstrates the advantage of EPC for Iris data where the data of three classes are more distinct that helps in search for rectangular rules. This advantage is a result of non-linear transformations in EPC in comparison with parallel coordinates.

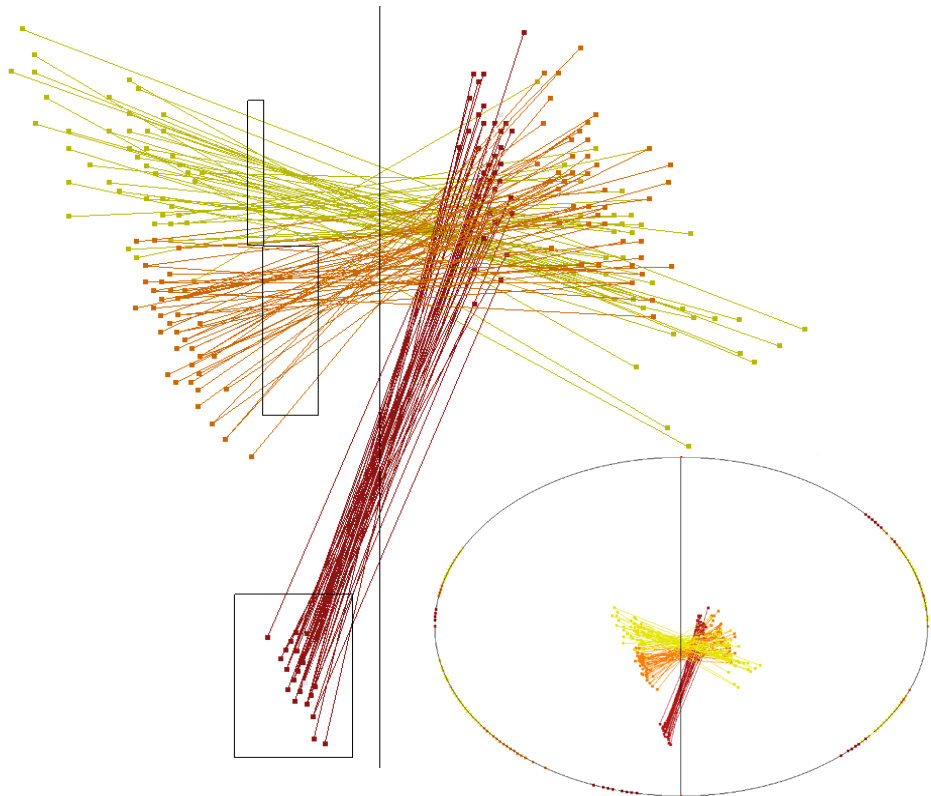

Fig. 3. Three Iris data classes classified by interactively created dominance rules (intersection based).

The rule **r₁** covers 52 cases (50 correct 2 misclassified cases), the rule **r₂** covers 48 cases (all 48 correct cases) and the rule **r₃** covers 50 cases (all 50 correct cases). These rules correctly classified 148 out of 150, i.e., 98.67%. Often each individual rule discovered in

EPC cover only a fraction of all cases of the dominant class. For such situations we use the **weighted precision** formula to compute the total precision of *k* rules

$$\sum_{i=1}^{k}(p_i c_i) / \sum_{i=1}^{k} c_i \qquad (1)$$

where $p_i$ is precision of rule **r**$_i$ and $c_i$ is the number of cases covered by **r**$_i$

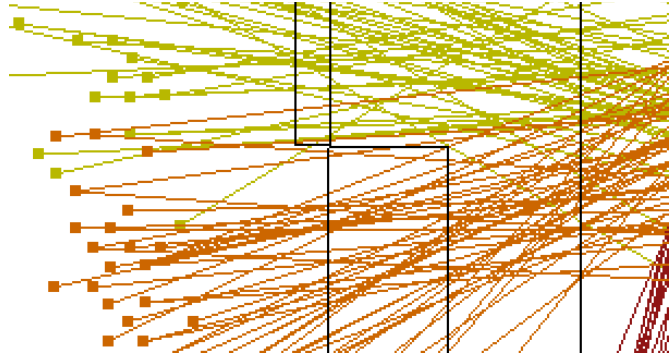

Fig. 4. Zoomed overlapped area from Fig. 3.

### 4.2 Visual verification of rules

Verification of discovered models is an important part of machine learning process. This section presents a simplified visual rule verification approach that has advantages over traditional k-fold cross validation (CV) is as we show with the Iris example.

A *random* split data to training and validation data to 10 folds is conducted in a common 10-fold CV approach to test the quality of prediction accuracy on the validation data. This random split does not guarantee that the worst-case split will be discovered and evaluated that is important for many applications with high cost or errors. In other words, cross validation can provide of the predictive model. EllipseVis allows to simplify or even eliminate cross validation.

Consider Iris data in Fig. 3 and randomly select 90% of lines to training and remaining 10% of them to validation data. Let these 90% of lines (training cases) cover the overlap area shown in Fig. 4, then we build rules **r**$_1$-**r**$_3$ as shows in Fig. 3 using these training data. The accuracy of these rules is 100% on validation data because these validation data are outside of the overlap area, confirming these rules.

In the opposite situation, when the training data do not include lines in the overlap area, but validation data include them, the discovered dominance rules (rectangles) can differ from **r**$_1$-**r**$_3$. Rule **r**$_1$ can be shifted lower or **r**$_2$ can be shifted higher and misclassify some validation cases. These cases are misclassified because the training data are not representative for these validation data. This is a **worst-case** cross validation split of data to training and validation when the training data are not representative for the validation data.

The visual EPC representation allows to find and see the worst split, and use this worst split, say, for selecting 15 Iris cases (10%) in the overlap area to a worst fold that include all overlap cases. A trivial **lower bound** of worst-case classification accuracy of this fold is 0 when all cases that are in this fold are misclassified. In contrast 9 other folds of 10-fold CV will likely be recognized with 100% accuracy by a well-designed and trained rules/models when considered as validation folds, because these folds are outside of the overlap area. Thus, the average 10-fold CV accuracy will be 90%. So, finding accurate rules such as **$r_1$-$r_3$** on the full dataset likely indicates that 10-fold CV result will be similar. Thus, instead of CV we can focus on finding visually the worst split and evaluate its accuracy on validation data [1]. It will likely be greater than 0 in general as it is the case for the iris data. Therefore, the 10-fold average will be greater than 90%. A more general and detailed treatment of the worst-case visual cross validation based on the Shannon function can be found in [9].

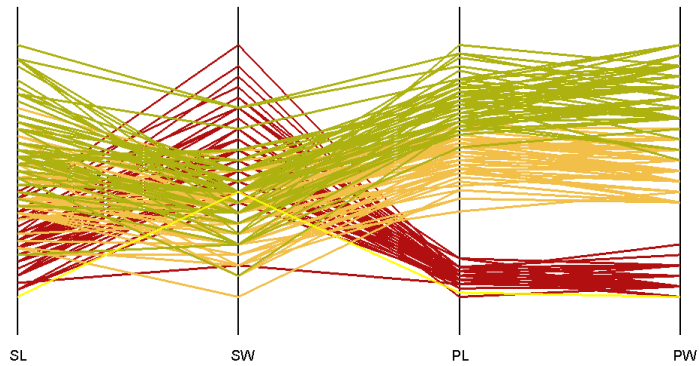

Fig 5. Iris data in parallel coordinates.

### 4.3 Experiment with Wisconsin breast cancer data

The goal of this experiment is testing EPC capabilities for the Wisconsin Breast Cancer (WBC) dataset [4] that consists of 683 full 9-D cases: 444 benign (red) and 239 malignant (green) cases as shown in Figs. 6-9 in EPC. To get an even number of coordinates we doubled coordinate $X_9$ creating $X_{10}$. Table 1 and Figs. 6-8 show the *automatically discovered* rules in the EllipseVis system. Together five simple rectangular rules cover **96.34%** of cases with weighted precision **95.13%**. Fig. 9 shows an example where interactive rule discovery is less efficient than automatic requiring two times more rules and larger rectangles.

TABLE 1. RESULTS OF AUTOMATIC WBC RULES DISCOVERY IN EPC (INTERCEPT-BASED).

| Rule | Class | Coverage/recall in class, % | Precision, % |
|---|---|---|---|
| $r_1$ | B | 64.18 | 98.59 |
| $r_2$ | B | 33.10 | 92.51 |
| $r_1$ or $r_2$ | **B** | **97.28** | |
| $r_3$ | M | 42.67 | 92.17 |
| $r_4$ | M | 37.23 | 92.13 |
| $r_5$ | M | 14.64 | 97.14 |
| $r_3$ or $r_4$ or $r_5$ | **M** | **94.54** | |
| **All rules** | **B, M** | **96.34** | **95.13** |

### 4.4. Experiment with multi-class Glass data

The 10-D multi-class Glass Identification dataset consists of 214 cases of 6 imbalanced classes of glass [4] used in criminal investigations by the U.S. Forensic Science Service. Fig. 10 visualizes all Glass classes in EPC. Below we show results of one class vs. all other classes.

**All other classes vs. class 5.** Table 2 shows all three rules discovered. They cover **87.06%** of cases of all other classes with weighted precision **98.29%**.

TABLE 2. 10-D GLASS RULES FOR CLASS 5 VS. ALL OTHERS (POINT-BASED).

| Rule | Class | Coverage/recall in class, % | Precision, % |
|---|---|---|---|
| $r_1$ | All but 5 | 37.31 | 100 |
| $r_2$ | All but 5 | 38.81 | 98.72 |
| $r_3$ | All but 5 | 10.95 | 90.91 |
| **All rules** | All but 5 | **87.06** | **98.29** |

**All other classes vs. class 6.** Fig. 11 shows the result of rule discovery to separate cases of all other classes from class 6. The EllipseVis system found three rectangles, $R_1$-$R_3$. and respective rules $r_1$-$r_3$. Total all three rules **cover 99.51%** of cases with weighted **precision 95.59%** (see Table 3).

TABLE 3. 10-D GLASS RULES FOR CLASS 6 VS. ALL OTHERS (POINT-BASED).

| Rule | Class | Coverage/recall in class, % | Precision, % |
|---|---|---|---|
| $r_1$ | All but 6 | 50.24 | 99.03 |
| $r_2$ | All but 6 | 18.05 | 94.59 |
| $r_3$ | All but 6 | 31.22 | 90.63 |
| **All rules** | All but 6 | **99.51** | **95.59** |

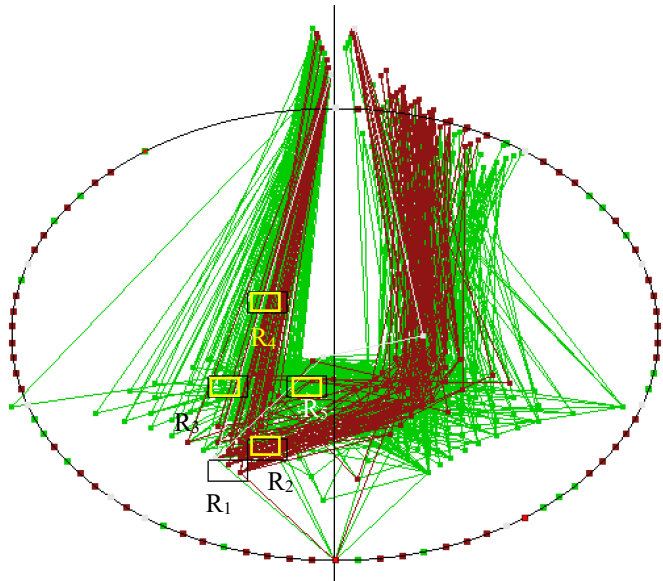

Fig. 6. WBC cases covered by automatically discovered rules $r_1$- $r_5$ with rectangles $R_1$- $R_5$ :
658 out of 683 cases (96.34%) with 95.13% precsion.

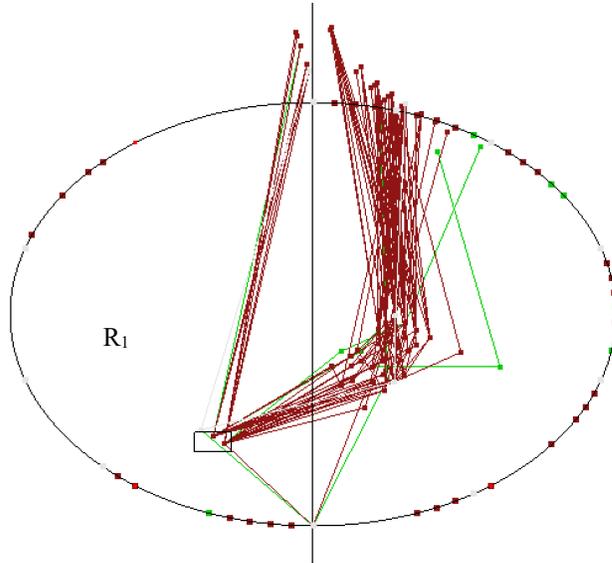

Fig. 7. Cases that satisfy rule **r**$_1$ based on the rectangle R$_1$ that covers 285 cases out of 444 cases of this class (64.18%) with 98.59% precision.

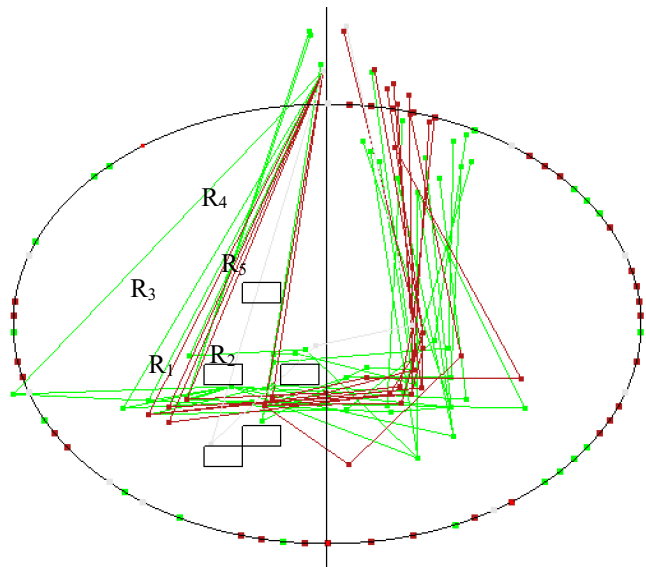

Fig. 8. Automatic rule discovery: remaining cases that do not satisfy rules **r**$_1$ - **r**$_5$ based on the rectangles R$_1$- R$_5$ with 96.34% of total coverage/recall.

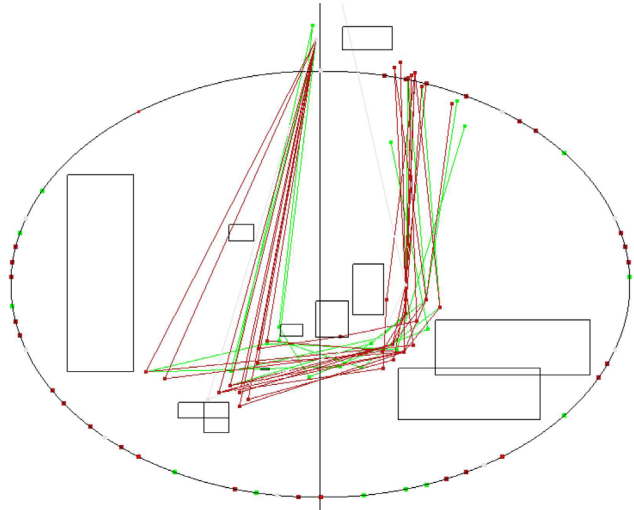

Fig. 9. Interactive "manual" rule discovery: remaining cases that do not satisfy discovered rules based on the shown rectangles with 96.63% of total coverage/recall with two times more rules.

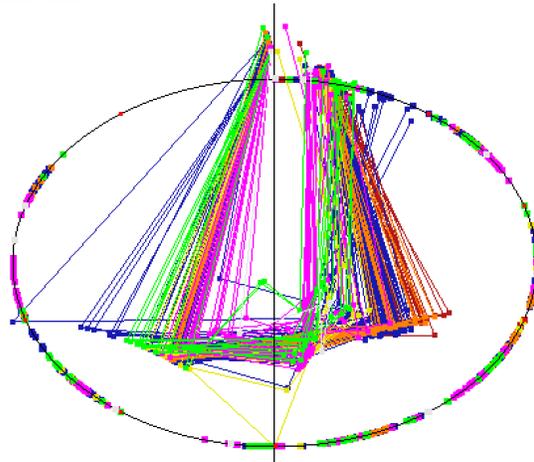

Fig.10. All Glass data of six classes in EPC

**All other classes vs. class 7.** Total all three rules **cover 87.57%** of cases with total **precision 97.31%** weighted by coverage (see Table 4). In addition, EllipseVis discovered rule $r_1$ of **class 7 vs. all others** that covered 79.31% of all cases of class 7 with 91.30 % precision (see Table 5). These rules are intercept-based and respectively, capture the non-linear relations between 4 attributes, which form a line that intersects the dominance rectangle), while point-based rules capture them between 2 attributes.

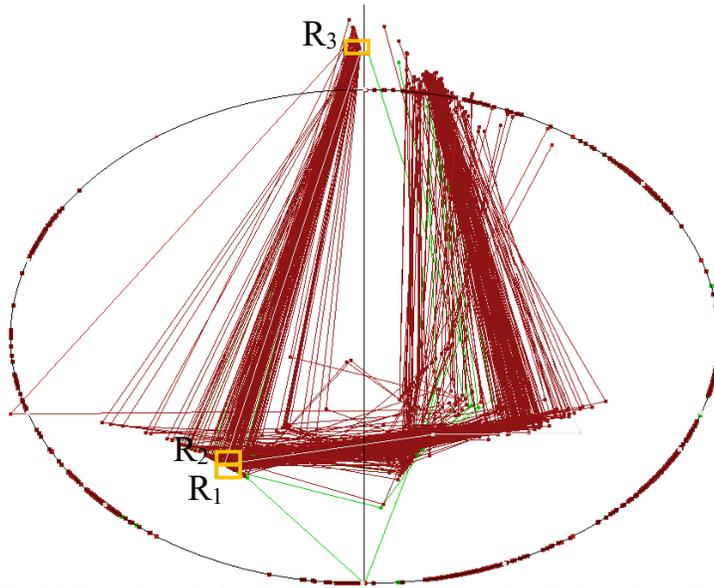

Fig. 11. Extracting rule for Glass: all other classes vs. class 6. Cases that satisfy extracted 3 rules $r_1$-$r_3$ based on the rectangles $R_1$- $R_3$.

TABLE 4. GLASS RULES FOR CLASS 7 VS. ALL OTHERS (POINT- BASED).

| Rule | Class | Coverage/recall in class, % | Precision, % |
|---|---|---|---|
| $r_1$ | All but 7 | 14.05 | 96.15 |
| $r_2$ | All but 7 | 54.59 | 100.00 |
| $r_3$ | All but 7 | 18.92 | 91.43 |
| **All rules** | All but 7 | **87.57** | **97.53** |

TABLE 5. GLASS RULES FOR CLASS 7 VS. ALL OTHERS (INTERSECT-BASED).

| Rule | Class | Coverage/recall in class, % | Precision, % |
|---|---|---|---|
| $r_1$ | 7 | **79.31%** | **91.30** |

### 4.5. Experiment with Car data

Figs. 12, 13 and Table 6 show the results for Car data with total **coverage of 91.24% and 100% precision.**

TABLE 6. CAR DATA RULE EXPERIMENTATION RESULTS FOR CLASS UNACC (POINT-BASED)

| Rule | Class | Coverage/recall in class, % | Precision, % |
|---|---|---|---|
| $r_1$ | UNACC | 15.87 | 100 |
| $r_2$ | UNACC | 15.87 | 100 |
| $r_3$ | UNACC | 15.87 | 100 |
| $r_4$ | UNACC | 23.80 | 100 |
| $r_5$ | UNACC | 7.93 | 100 |
| $r_6$ | UNACC | 3.97 | 100 |
| $r_7$ | UNACC | 3.97 | 100 |
| $r_8$ | UNACC | 3.97 | 100 |
| **All rules** | UNACC | 91.24 | 100 |

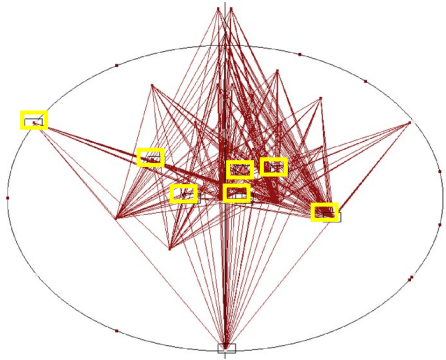
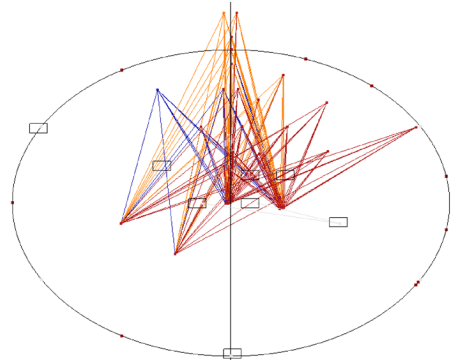

Fig. 12. Extracting rules for Car data: cases of class unacc (red) that satisfy extracted 8 rules $r_1$-$r_8$ based on the rectangles $R_1$ - $R_8$. (yellow)

Fig. 13. Remaining Car cases of three other classes that do not satisfy rules $r_1$ - $r_8$ based on the rectangles $R_1$- $R_8$.

### 4.6. Experiment with Ionosphere data

Tables 7, 8 and Figs. 14, 15, show results with Ionosphere data [4] for both intercept- and point-based rules. These rules cover both classes with similar coverage (**78.63, 71.51**) and precision (**91.37, 94.02**).

TABLE 7. 34-D IONOSPHERE DATA DOMINANCE RULE EXPERIMENTATION RESULTS (INTERSECT-BASED)

| Rule | Class | Coverage/recall in class, % | Precision, % |
|---|---|---|---|
| $r_1$ | b | 30.95 | 97.87 |
| $r_2$ | b | 34.13 | 90.70 |
| $r_1$ or $r_2$ | b | **65.08** | |
| $r_3$ | g | 18.22 | 90.24 |
| $r_4$ | g | 27.11 | 90.16 |
| $r_5$ | g | 40.89 | 90.22 |
| $r_3$ or $r_4$ or $r_5$ | g | **86.22** | |
| All rules | b,g | **78.63** | **91.37** |

TABLE 8. 34-D IONOSPHERE DATA DOMINANCE RULE EXPERIMENTATION RESULTS (POINT BASED)

| Rule | class | Coverage/recall in class, % | Precision, % |
|---|---|---|---|
| $r_1$ | b | 29.37 | 100.00 |
| $r_2$ | b | 30.16 | 97.37 |
| $r_1$ or $r_2$ | b | **59.53** | |
| $r_3$ | g | 54.67 | 90.24 |
| $r_4$ | g | 23.56 | 96.23 |
| $r_3$ or $r_4$ | g | **78.23** | |
| All rules | b,g | **71.51** | **94.02** |

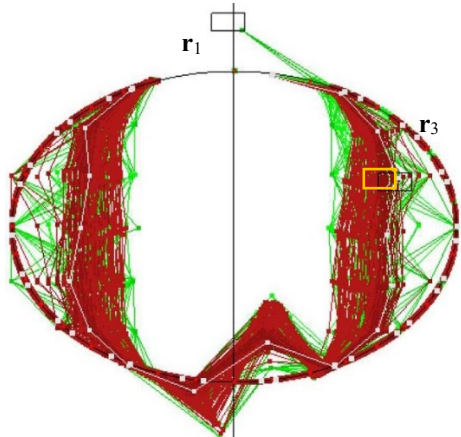 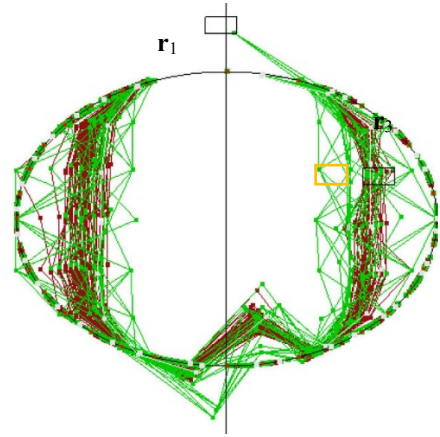

Fig 14. Ionosphere data with rules $r_1$ (green) and $r_3$ (red).

Fig 15. Data that satisfy Rule $r_1$ (green class) for Ionosphere with data with cases from red class that satisfy Rule $r_3$ from red class on the background.

*4.7. Experiment with Abalone Data*

Fig. 15 and Table 9 present the results for Abalone data [4] on full data and Table 10 shows in split 70%-30% to training and validation data for two classes (green and red).

TABLE 9. 8-D ABALONE DATA DOMINANCE RULE EXPERIMENTATION RESULTS (POINT BASED)

| Rule | class | Coverage/recall in class, % | Precision, % |
|---|---|---|---|
| $r_1$ | 1 | 45.12 | 92.83 |

TABLE 10. 8-D ABALONE DATA 70%:30% SPLIT (POINT BASED)

| Rule | Class | Training | | Validation | |
|---|---|---|---|---|---|
| | | Coverage, % | Precision, % | Coverage, % | Validation, % |
| $r_1$ | 1 | 12.12 | 91.18 | 12.59 | 94.12 |
| $r_2$ | 1 | 11.28 | 90.08 | 12.59 | 88.24 |

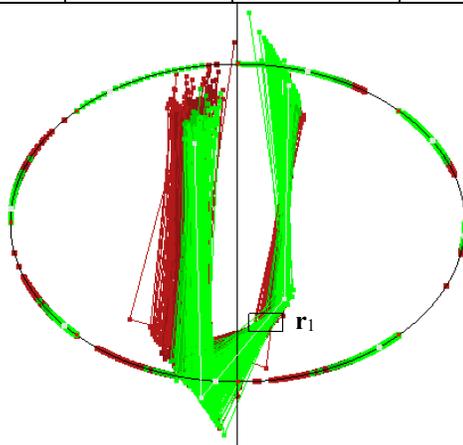

Fig 16. Abalone data with rule $r_1$.

*4.8. Experiment with Skin segmentation data*

This experiment is to explore abilities to build EPC rectangular rules on a large skin segmentation dataset [4], which contain 245,000 cases of three dimensions and two classes. Below we present discovered rules for these data based on two methods: (1) the number of points within the rectangle and (2) the number of lines that intersects the rectangle.

We use the following notation: skin class (class 1, red) and non-skin class (class 2, green). In Figs. 17-20, the darker green color shows the cases of the green class which have already been included in previous rules. The current version of EPC requires an even number of attributes. Therefore, a new attribute $x_4$ is generated to be equal to 1.0 for all cases on the normalized [0,1] scale. The rendering 245,000 cases in EPC is relatively slow in the current implementation: approximately 20 seconds, compared to less than one second for much smaller WBC data with less than 700 cases. Approximately 50,000 cases are skin class.

The discovered **point-based rules and intersect line-based rules** are presented, respectively, in Table 10 and Figs. 17-18, and Table 11 and Figs. 19-20. Fig. 17 and 19 shows non-skin cases (green) in front of the skin cases (red) with automatically and sequentially discovered rectangles for green class rules. Figs. 18 and 20 show the opposite order of the green and red cases.

While the precision or rules for Skin segmentation data shown in Tables 10 and 11 is quite high (94.62%, 97.78%), the coverage is relatively low (61.4%, 42%). It is likely related to the *large size* of this dataset (245,000 cases) where the data are hardly can be *homogeneous* to be covered by few rectangles, which cover over 10% of the data each.

The *higher dimensions* (Ionosphere data) is another likely reason why that data are non-homogenous requiring more rectangles. *Multiple imbalanced classes* with some extreme classes which contain a single case (Abalone data) is another possible reason why that data are non-homogenous requiring more rectangles too. These issues suggest areas of further development and improvement for EPC. We explore generalizations of EPC in Section 6 to address these issues in the future work.

Table 10. Point-based Rules.

| Rule | Class | Coverage/recall, % | Coverage/recall, # | Precision % |
|---|---|---|---|---|
| $r_1$ | 2 | 10.56 | 25878 | 100 |
| $r_2$ | 2 | 10.83 | 26545 | 99.94 |
| $r_3$ | 2 | 32.32 | 79199 | 90.9 |
| $r_4$ | 2 | 7.69 | 18838 | 95.34 |
| All Rules | 2 | 61.4 | 150460 | **94.62*** |

*weighted precision.

Table 11. Intersect-based Rules.

| Rule | Class | Coverage/recall, % | Coverage/recall, # | Precision % |
|---|---|---|---|---|
| $r_1$ | 2 | 17.64 | 43230 | 100 |
| $r_2$ | 2 | 11.27 | 27622 | 100 |
| $r_3$ | 2 | 13.09 | 32093 | 92.87 |
| All Rules | 2 | 42.00 | 102945 | **97.78*** |

*weighted precision.

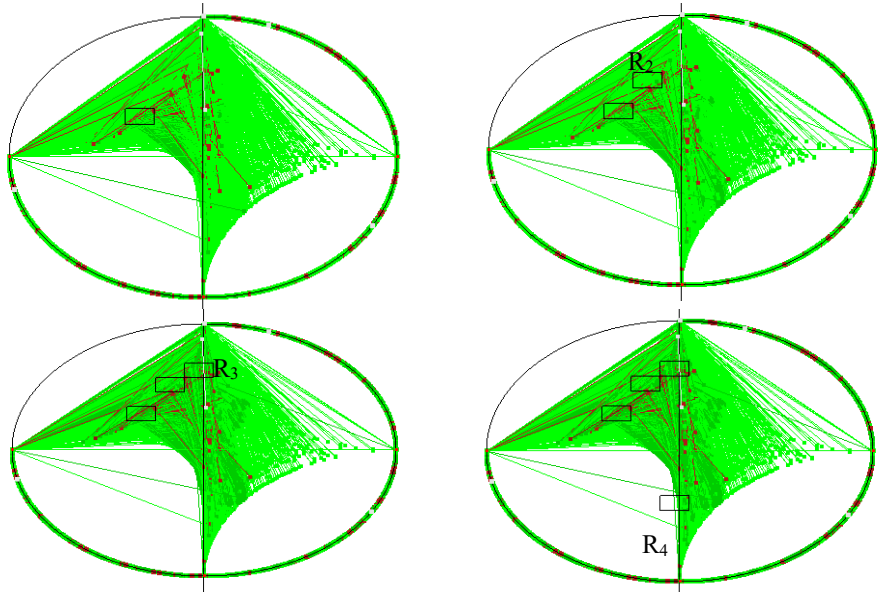

Fig. 17. Skin segmentation data in EPC with point-based rules $r_1$-$r_4$ and green cases in front of red cases

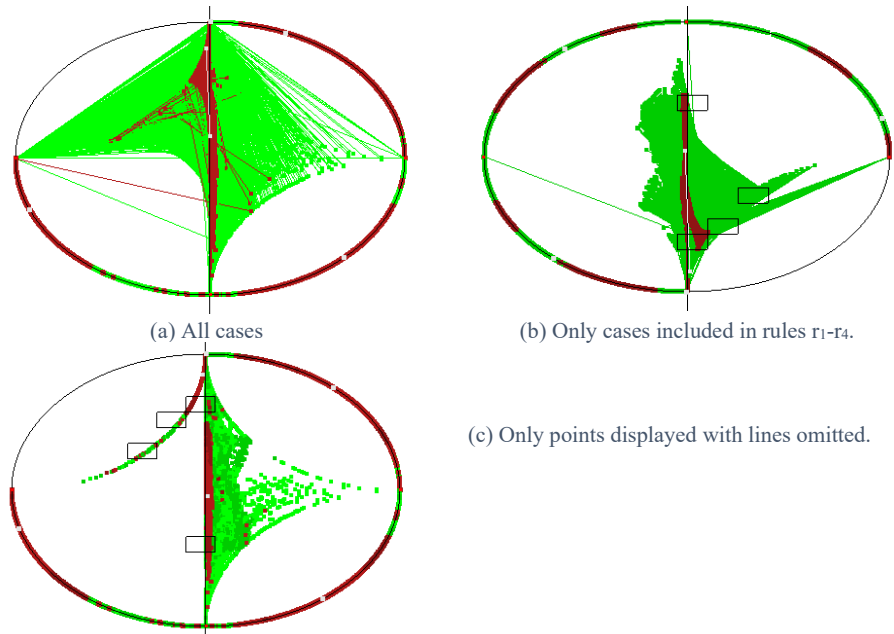

(a) All cases

(b) Only cases included in rules $r_1$-$r_4$.

(c) Only points displayed with lines omitted.

Fig. 18. Skin segmentation data in EPC with point-based rules $r_1$-$r_4$ and red cases in front of green cases.

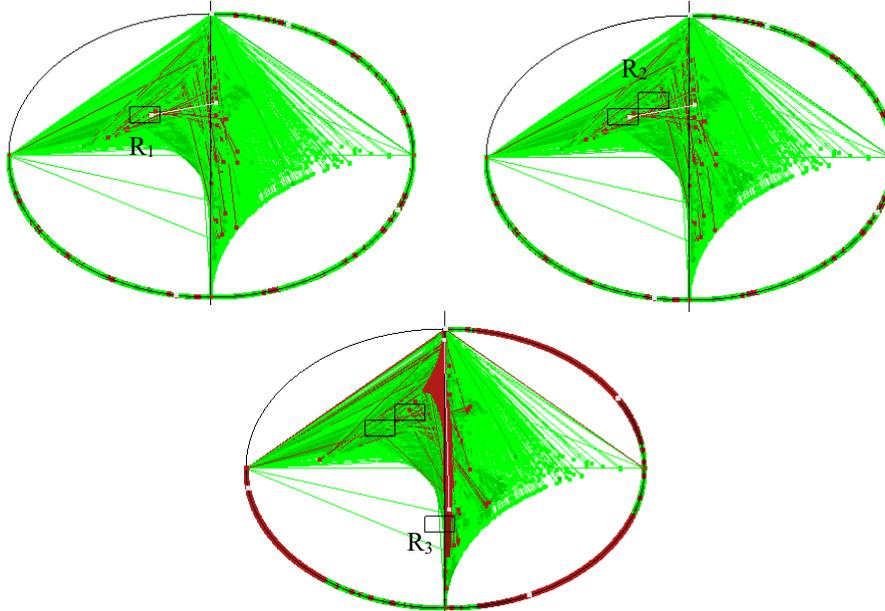

Fig, 19. Skin segmentation data in EPC with intersect-based rules $r_1$-$r_3$ and green cases in front of red cases.

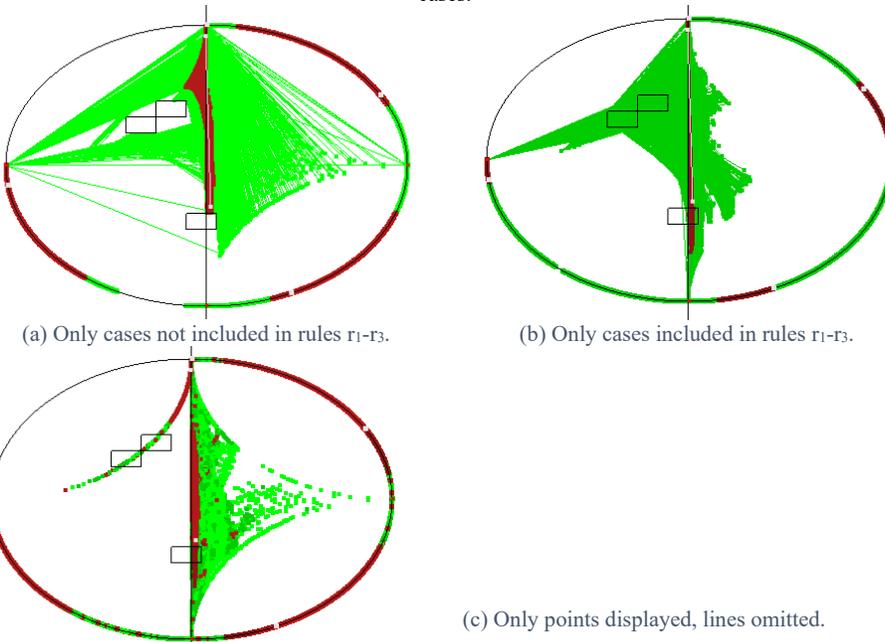

(a) Only cases not included in rules $r_1$-$r_3$.   (b) Only cases included in rules $r_1$-$r_3$.

(c) Only points displayed, lines omitted.

Fig, 20. Skin segmentation data in EPC with intersect-based rules $r_1$-$r_3$ and red cases in front of green cases.

## 5. EXPERIMENT WITH SYNTHETIC DATA

The goal of this experiment is to explore the abilities of EPC to represent data as simple easy recognizable shapes like straight horizontal lines, rectangles, and others by using synthetic data.

**Experiment S1: complex dependence**. This experiment was conducted with a **set A** of 9 synthetic 8-D data points $\mathbf{x}_1$=(0.9, 0.1, 0.9, 0.1,…), $\mathbf{x}_2$ =(0.8, 0.2, 0.8, 0.2,…), …., $\mathbf{x}_9$= (0.1, 0.9, 0.1,0.9,…), which have complex dependencies within each 8-D points and between points. The non-linear dependence within each 8-D point is:

$$\text{if } j \text{ is odd then } x_{i+1,j} = k, \text{ else } x_{i+1,j} = 1-k.$$

and non-linear dependence between consecutive 8-points $\mathbf{x}_i$ and $\mathbf{x}_{i+1}$ is:

$$\text{if } j \text{ is odd then } x_{i+1,j} = x_{i,j} - 0.1, \text{ else } x_{i+1,j} = x_{i,j} + 0.1.$$

These data are shown in EPC and parallel coordinates Fig. 21 where it is visible that their pattern is more complex in parallel coordinates than in EPC.

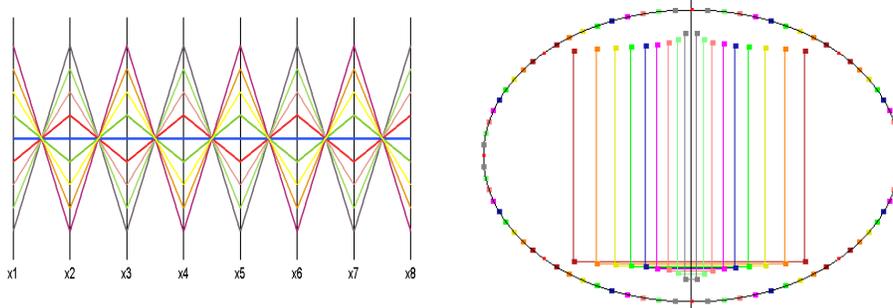

Fig. 21. 8-D synthetic dataset A in parallel coordinates and EPC.

**Experiment S2: linear dependence**. For this experiment, we generated a **set B** of nine 4-D points $\mathbf{x}_1$=(0.1, 0.1, 0.1, 0.1), $\mathbf{x}_2$=(0.2, 0.2, 0.2, 0.2), …, $\mathbf{x}_9$=(0.9, 0.9, 0.9, 0.9) with equal values within each 4-D point. Points $\mathbf{x}_i$=($x_{i1}, x_{i2}, x_{i3}, x_{i4}$) have simple linear dependences: $x_{ij} = x_{ik}$ and $x_{i+1,j} = x_{i,j} + 0.1$.

Fig. 22 shows these 4-D points as lines of different colors in EPC and parallel coordinates. In parallel coordinates, 9 parallel lines show these 4-D points. These lines do not overlap and respectively simpler for visual analysis than EPC. This is expected because each of these 4-D points satisfy simple linear dependences, while EPC are designed for non-linear dependencies.

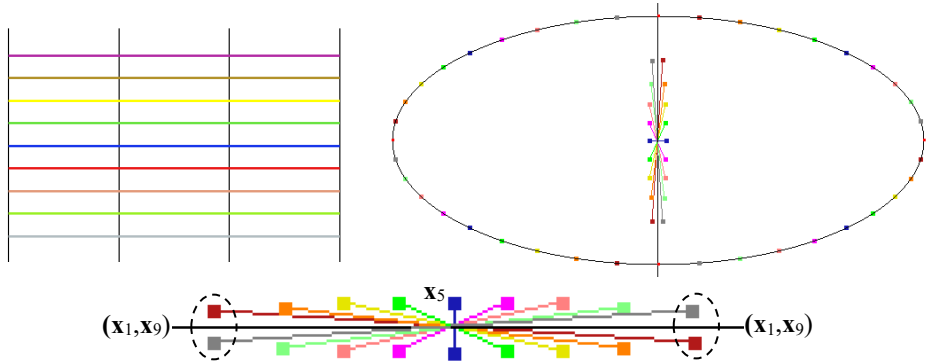

Fig. 22. Experiment S2 with data in parallel coordinates, EPC and zoomed and rotated EPC.

In EPC, the dark blue 4-D point $x_5$ in the middle is the shortest line perpendicular to M axis and $x_4$ and $x_6$ are next to it and $x_1$ and $x_9$ are far away from them. Similarly, lines for these points are located close or far away in parallel coordinates. Thus, both EPC and parallel coordinates capture similarities and differences by putting similar 4-D points next to each other and different far away in the visualization.

There is also a difference between EPC and parallel coordinates that is especially visible when $x_1$ and $x_9$ belong to the same class and we would like to have a visualization where they are close to each other in this visualization. Parallel coordinates do not do this. In Fig. 22, they are far away in parallel coordinates, while in EPC $x_1$ and $x_9$ are next to each other. Thus, this example shows that EPC allows visualizing cases of **non-compact classes** next to each other in contrast with parallel coordinates.

Since the graph is dependent on the central ellipse, changing the horizontal and/or vertical dimensions of this ellipse results in a stretched/shrunk graph. This helps to make the data points and lines more distinct in the respective axis being stretched, i.e. increased, and conversely compresses data for a decreased axis. Increased data point distinction within the ellipse is currently done by zooming in using the camera controls; effectively increasing both axes' dimensions.

**Experiment S3: complex dependence.** For this experiment, we generated a **set C** of nine 4-D points $x_1$=(0.1, 0.9, 0,1, 0.9), $x_2$=(0.2, 0.8, 0.2, 0.8), $x_3$=(0.3, 0.7, 0.3, 0.7), $x_4$=(0.4, 0.6, 0.4, 0.6), $x_5$=(0.5, 0.5, 0,5, 0.5), …, $x_9$=(0.9, 0.1, 0,9, 0.1). This is an inverse dependence within each 4-D point. The relation within and between these 4-D points are more complex than in the set B. Fig. 23 shows these data in EPC and parallel coordinates  The importance of this example is in the fact that all these 4-D points are located in the straight horizontal line producing a simple preattantive pattern. This is beneficial if all of them belong to a single class. In contrast in parallel coordinates on the right in Fig. 23 thue do not form a simple compact patter, but cover almost the whole area.

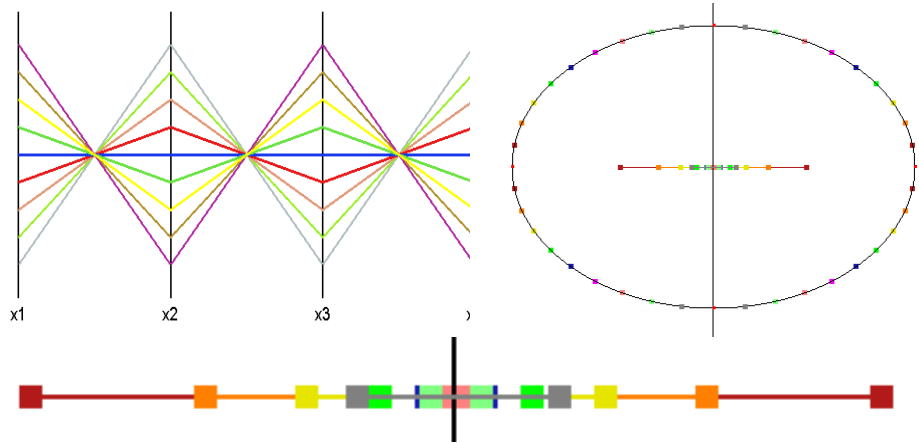

Fig. 23. Experiment S3 with data in parallel coordinates, EPC and zoomed EPC.

**Experiment S4: complex dependence**. In this experiment we visualized seven 4-D points with the following property: $(x_1, x_2, x_3, x_4) = (x_1, x_2, 1-x_1, 1-x_2)$. These points shown in Fig. 24 are (1/8, 1/8, 7/8, 7/8) (light green), (2/8, 2/8, 6/8, 6/8) (brown), (3/8, 3/8, 5/8, 5/8) (magenta), (4/8, 4/8, 4/8, 4/8) (blue),(5/8, 5/8, 3/8, 3/8) (darker green), (6/8, 6/8, 2/8, 2/8) (yellow) and (7/8, 7/8, 1/8, 1/8) (red).

In contrast with the Fig. 23 all these 4-D points are located vertically in EPC. If these points represent one class and points in Fig. 23 represent another class, then we easily see the difference between classes -- one is horizontal another one is vertical overlap that can be observed quickly pre-attentively.

Fig. 6 also shows this data in parallel coordinates. The patterns in in parallel coordinates are more complex. Each 4-D point is represented as a polyline that consists of 4 points and three segments that connect them while in EPC each 4-D point requires two points and one segment to connect them. Next, the polylines in parallel coordinates in this figure are not straight lines and therefore cannot be observed pre-attentively.

In addition, these polylines cross each other cluttering the display. In contrast these data in EPC in Fig. 24 are simple straight horizontal lines that do not overlap that can be observed pre-attentively.

Comparing these data in parallel coordinates in Figs. 23 and 24 allows to see the difference clearly when these two images are put side-by-side. However, when all data of both classes are in the single parallel coordinates the multiple lines cross each other and the difference between classes is muted. In contrast in EPC the lines of two classes overlap only in single the middle point.

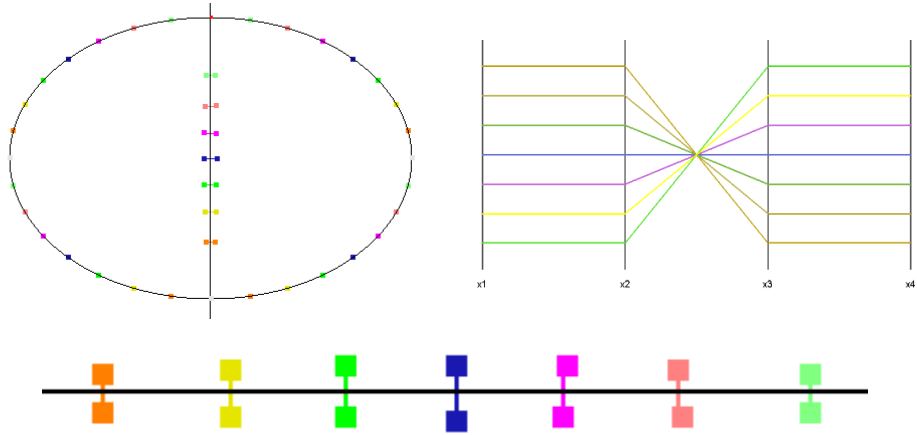

Fig. 24. Experiment S4 with data in parallel coordinates, EPC and zoomed and rotated EPC.

**Algorithm.** Example above show us that simple visual patterns are possible in EPC for the data that are quite complex in other visualizations. The next question is how to find n-D points that have a given simple visual pattern in EPC like a short straight line, a rectangle, or others.

The steps of the algorithm are as follows for a horizontal line (see Fig. 25 for illustration):
(1) Select a horizontal black line in EPC;
(2) Pick up a point $P_1$ on the horizontal black line;
(3) Shift the red right side ellipse to reach this point $P_1$;
(4) Shift the blue right ellipse to the same point $P_1$;
(5) Shift the blue left ellipse to touch the right blue ellipse;
(6) Shift the red left ellipse to touch the right red ellipse;
(7) Find the point where left red and blue ellipses cross each other and mark this point as $P_2$ (now both $P_1$ and $P_2$ are on the black line);
(8) Draw an arrow from $P_1$ to $P_2$;
(9) Find the points where blue and red ellipses cross the main ellipse
(10) Mark these points by $x_1, x_2, x_3, x_4$ to indicate that they represent values of $x_1, x_2, x_3, x_4$ of the 4-D point **x** that is on the black line.
(11) Repeat 1-10 for any other point on the back line.

Finding values in (9) requires taking the equation of the main ellipse and the equation of respective side ellipse and solving the system of these equations to find a crossing point that belongs to both. A general property of all 4-D points that are on the horizontal black line is that that $x_1=x_3$, and $x_2=x_4$. In other words, let **a** be a 4-D point on this line then all other points can be expresses as follows: $\mathbf{y} = (a_1+e_1, a_2+e_2, a_3+e_1, a_4+e_2)$, where $e_1$ and $e_2$ are function of the shifts of **y** relative to **a**.

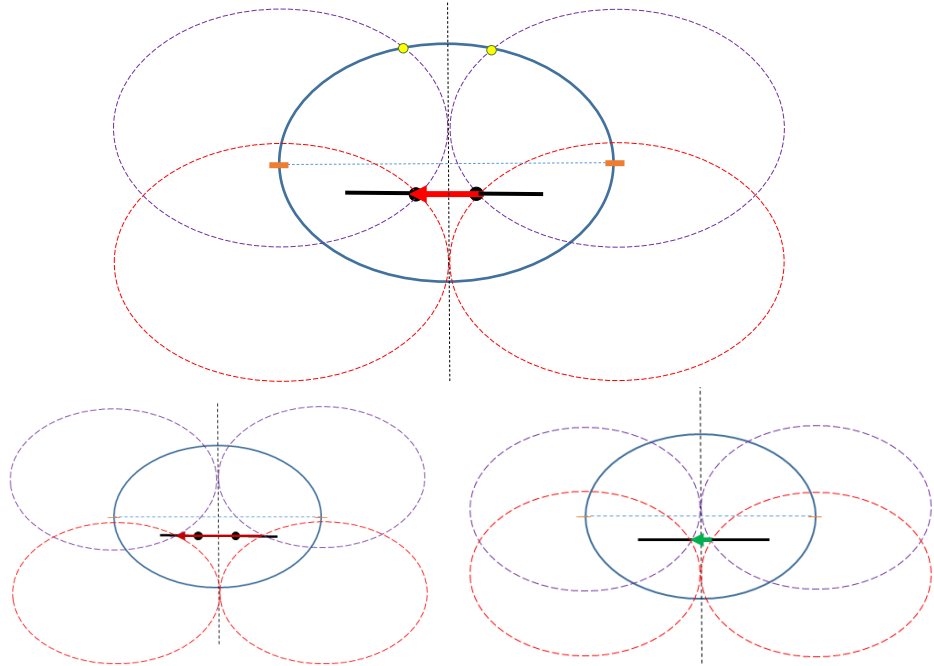

Fig. 25. Three 4-D points mapped to the back line as arrows in EPC

6. GENERALIZATION OF EPC

This section presents two generalizations of elliptic paired coordinates: the **dynamic elliptic paired coordinates (DEPC)** and **weights of coordinates**. The first one allows to represent more complex relations and the second one allows making selected attributes more prominent that can reflect their importance.

*6.1. Dynamic elliptic paired coordinates*

In DEPC the location of each next point $x_{i+1}$ **depends** on the location of the previous point $x_i$. DEPC differs from EPC that is *static*, i.e., the location of points $x_i$ does not depend on the location of the other points. Fig. 1 shows a 4-D point $\mathbf{x} = (x_1,x_2,x_3,x_4) = (0.3, 0.4, 0.2, 0.6)$ in the dynamic elliptic paired coordinates. in Fig. 26 the value of $x_1$ is located at 0.3 on the blue central ellipse. The value of $x_2$ is located at 0.3+0.4=0.7 on the same blue ellipse starting from the origin, the value of $x_3$ is located at 0.7+0.2=0.9, and the value of $x_4$ is located at 0.9+0.6=1.5, i.e., each next coordinate value starts at the location of the previous point. Fig.27 shows the same 4-D points in the static EPC.

The difference from the static EPC is that we do not need to create separate sections for each attribute on the EPC central ellipse, but start from the *common origin* at the top of the central ellipse (the grey dot in Fig. 26) and add value for the next attribute to the location of the previous one. Then the process of construction of points $P_1$ and $P_2$ is the same as in the static EPC. The advantages of DEPC is that it allows discovering more complex non-linear relations than EPC.

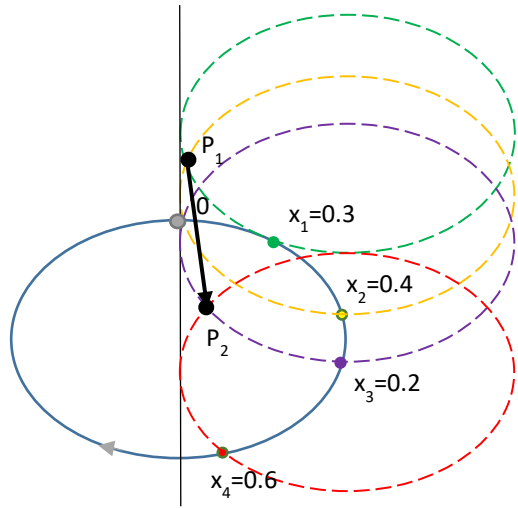

Fig. 26. 4-D point $\mathbf{x} = (0.3, 0.4, 0.2, 0.6)$ in dynamic EPC.

There are several options to locate side ellipses relative to the point $x_i$ on the central ellipse: (1) the center of the side ellipse is above the point $x_i$ (ellipse goes up), or (2) the center of the side ellipse is below the point A (ellipse goes down). Next, (1) and (2) can be in different combination as shows in Fig 28. Fig 28a shows the case when the red ellipses go down for values $x_2$ and $x_4$, but the blue ellipses go up for values $x_1$ and $x_3$. Fig. 3b shows the case when all ellipses go down and Fig. 28c shows the case when all ellipses go up. Respectively points $P_1$ and $P_2$ are differently located relative to the central ellipse as shows in these figures. Both points $P_1$ and $P_2$ are in the central ellipse in Fig. 3a, $P_2$ is outside in Fig. 3b and $P_1$ is outside in Fig. 3c. In Fig. 1 all side ellipses go up (centers above the $x_i$ point), and in Fig. 2 red ellipses go down, and blue ellipses go up with $P_1$, $P_2$ within the central ellipse, while in Fig.1 only $P_2$ is in the central ellipse.

### 6.1. EPC with odd number of coordinates and alternative side ellipses

EPC coordinates have been defined in section 2. Below we present alternative definitions. Such alternative definitions expand the abilities to represent n-D data in EPC differently opening an opportunity to find representation that will be most appropriate for particular n-D data and specific machine learning task on this data.

The EPC defined in Section 2 require an even number of coordinates. For the odd number of coordinates, we artificially add another coordinate either by copying one of coordinates or set up an additional coordinate be equal to a constant for all cases.

However, there is a way to visualize in EPC the odd number of coordinates as Fig. 29 illustrates for the 3-D point $\mathbf{x} = (x_1, x_2, x_3) = (0.3, 0.25, 0.6)$. Here ellipses for $x_1$ and $x_3$ are build relative to the horizontal line N (right (yellow) and left (green) side ellipses), but ellipse for $x_2$ is built relative to vertical line M (right (red) ellipse).

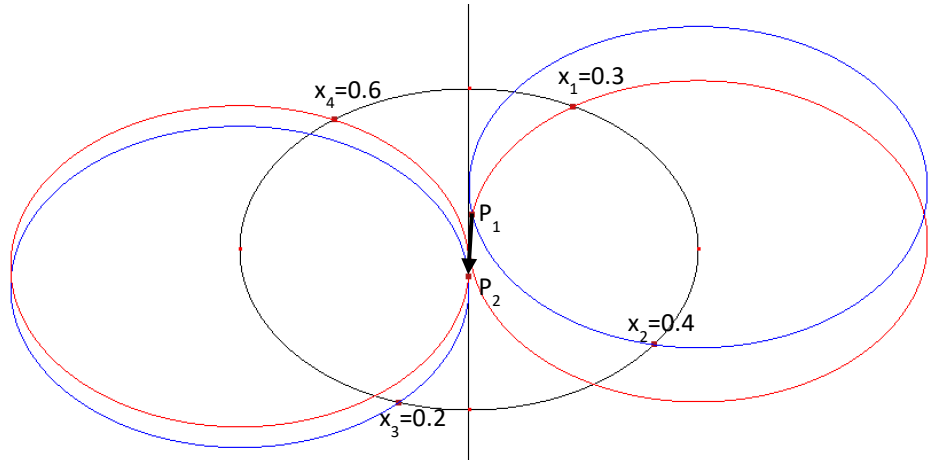

Fig. 27. 4-D point **x** = (0.3, 0.4, 0.2, 0.6) in static EPC.

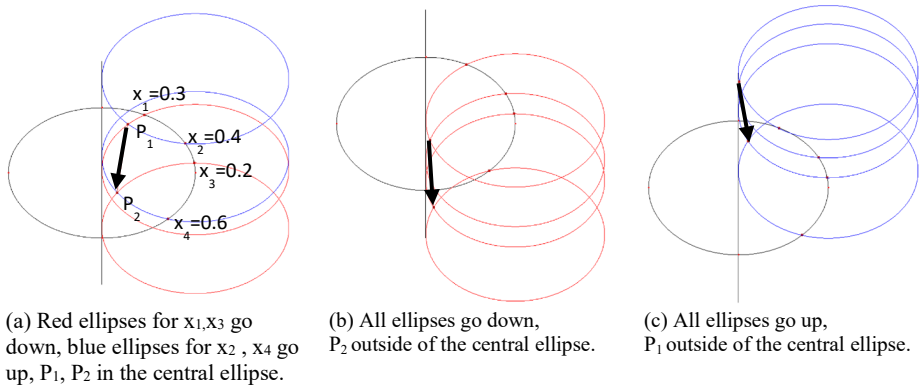

(a) Red ellipses for $x_1, x_3$ go down, blue ellipses for $x_2$, $x_4$ go up, $P_1$, $P_2$ in the central ellipse.

(b) All ellipses go down, $P_2$ outside of the central ellipse.

(c) All ellipses go up, $P_1$ outside of the central ellipse.

Fig. 28. Three version of Dynamic EPC.

This idea of using a mixture of top, bottom, left and right ellipses is expandable to the situations with any number of coordinates odd or even. Some of these ellipses can be used for some coordinates while others for other coordinates in multiple possible combinations. The number of these combinations is quite large resulting in a wide variety of visualizations with an opportunity to optimize the visualization for both human perception and machine learning.

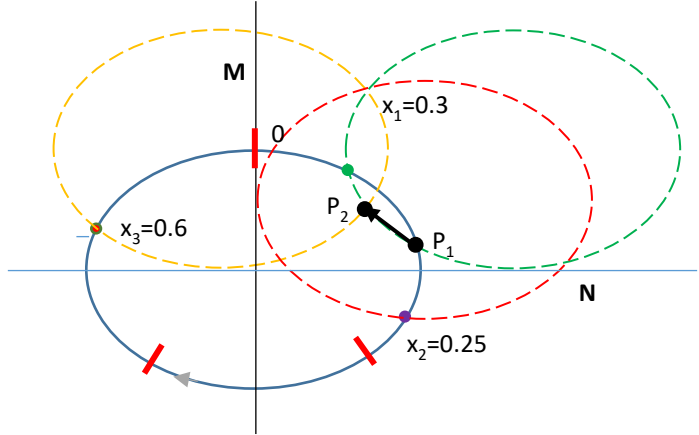

Fig. 29. EPC with mixture of right and top side ellipses.

*6.2. EPC with weighs of coordinates*

The introduction of weights of the attributes $x_i$ allows to build EPC visual representations where some attributes will be more prominent than others. A *base option* is multiplying every value of $x_i$ by its weigh $w_i$ with using $w_i x_i$ in EPC instead of original $x_i$ to build a visual representation of n-D point **x**. This option can fail when the value of $w_i x_i$, will be out of the range of the ellipse segment assigned to the coordinate $x_i$. It can also exceed the whole ellipse. This can happen when the values of weights are not controlled.

In a **controlled static option**, *the length of segments* of the ellipse associated with each coordinate $x_i$ are *proportional* to its weight $w_i$. For example, consider, four segments for attributes $x_1$-$x_4$ with respective weigh $w_i$ as 4, 2, 6, and 5. Then 4/17, 2/17, 6/17 and 5/17 will be fractions of the ellipse circumference assigned to $x_1$-$x_4$, respectively. If $x_1$=0.3 then it's location $x_{iw}$ will be 0.3*4/17 fraction of the ellipse circumference with a general formula:

$$x_{iw} = w_i x_i / (\sum_{i=1}^{n} w_i) \quad (1)$$

A **controlled dynamic option** is using the values $x_{iw}$ from (1) instead of $x_i$ in DEPC without creating separate segments for each coordinate.

**Assigning and optimizing weights**. Weights can be assigned by a user interactively or can be optimized by the program in the following way by first making all weights $w_i$ =1 and then adjusting them with steps $\delta(w_i)$. After each adjustment of the set of weights, the program computes *accuracy* of classification and other quality indicators on the training data in search for the best set of weights that maximize the selected quality indicators. The adjustments can be conducted randomly, adaptively, by using genetic algorithms and others. The *compactness* of location of cases each class and far away from case of other classes is one of such quality indicators.

*6.3. Incompact machine learning tasks*

The two generalizations presented above open the opportunity for solving **incompact machine learning** tasks – tasks with cases of the same class that are far away from each other in the feature space. A traditional assumption in machine learning is the **compactness hypothesi**s that points of one class are located of next each other in the feature space, in other words, similar real-world objects have to be close in the feature space [8]. This is a classical assumption in *k*-nearest neighbor algorithm, Fisher Linear Discrimination Function (LDF) and other ML methods.

Experiment S2 in Section 5 had shown an example of n-D points $x_1$ and $x_9$ that are far away from each other in the feature space are close to each other in EPC space. Thus, EPC has capabilities needed for incompact ML. The use of *weights* for attributes can enhance these capabilities. Thus, we want to find area $A_1$ in the EPC where the cases of a one class will be compactly located in spite likely being far away in the original feature space and cases another class will concentrate in another area $A_2$. Similarly, as we had shown in Figs. 18, 5 and 6 in section 5.

While we have these positive these examples, it is desirable to prove mathematically that EPC and DEPC have enough power for learning such distinct areas $A_1$ and $A_2$ with optimization of weights for the data with a wide range of properties. For instance, this can be data where each class consists of the cases that belong to multidimensional normal distributions located far away from each other.

The rectangles we discovered for real data in section 4 partially satisfy this goal. It is not a single rectangle for each class but several rectangles. Next, these rectangles contain only a part of each n-D point. In contrast the compactness hypothesis expects that full n-D points will be in some local area of the feature space. In other words, the compactness hypothesis assumes a space where each n-D point is a single point in this n-D space. In contracts in EPC each n-D point is graph in 2-D EPC space. Thus the compactness hypothesis in EPC space must differ from the traditional formulation requiring only a part of the graphs of cases of a class localized in the rectangle. Next the minimization of the number of rectangles is another important task, while reaching a single rectangle per class can be impossible. Also, these rectangles may not cover all cases of the class as we have seen in Section 4. All these rectangles are not in the predefined locations, because they discovered with fixed weights without optimization of weights. Optimization of weighs can make the compactness hypothesis richer allowing rectangles to be in the predefined locations by adjusting weighs combined with dynamic EPC.

## 7. SUMMARY AND CONCLUSION

The results of all experiments with EPC using the EllipseVis are summarized in Table 12 showing precision of rules from 91% to 100% with recall from 42% to 100% and 1-8 rules per experiment. They produced a small number of simple, visual rules which show cases clearly where a given class dominates. The end users and domain experts who are not machine learning experts can discover these rules themselves using EllipseVis doing end-user self-service. The visual process does not require mathematical knowledge of ML algorithms from users to produce these visual rules in the EllipseVis system.

TABLE 12. DOMINANCE RULE EXPERIMENTATION RESULTS

| Experiment | n-D | Classes | Rules | Recall, % | Precision, % |
|---|---|---|---|---|---|
| Iris | 4 | 3 | 3 | 100 | 98.66 |
| Cancer | 9 | 2 | 5 | 96.33 | 95.13 |
| Glass 1 | 10 | 2[a] | 3 | 87.06 | 98.29 |
| Glass 2 | 10 | 2[a] | 3 | 99.51 | 95.59 |
| Glass 3 | 10 | 2[a] | 3 | 87.57 | 97.53 |
| Glass 4 | 10 | 2[a] | 1 | 79.31 | 91.30 |
| Car | 6 | 4 | 8 | 91.24 | 100.00 |
| Ionosphere 1 | 34 | 2 | 5 | 78.63 | 91.37 |
| Ionosphere 2 | 34 | 2 | 4 | 71.51 | 94.02 |
| Abalone | 8 | 2 | 1 | 45.12 | 92.83 |
| Skin 1 | 4 | 2 | 4 | 61.40 | 94.62 |
| Skin 2 | 4 | 2 | 3 | 42.00 | 97.78 |

[a]. One class vs. all other classes.

In addtion, the accuracy of the presented results are competitive with reported in [1,5] and exceed some of the other published resutls, while the goal is not competing with them in accuracy, but showing opprtunity for the the end users and domain experts to discover understandable rules themselves as *self-service* without programming and studying mathematical intricacies of ML algorithms. Another brenefit for the end-users is ablities to Find the visually the worst split of the data for training and validation and evaluating the accuracy of classification on this split getting the worst case estimate for algorithm accuracis it is demonstrated with the Iris data. Several options of the further development are outlined in section 6 that include generalization of elliptic paired coordinates to the dynamic elliptic paired coordinates, incorporating and optimizing weights of these coordinates and intorducing incompact machine learning tasks. As development and evaluation of EllipseVis and EPC continues, we will be able to gain a better understanding of EPC's capabilities to construct a data visualization for visual knowledge discovery.